\documentclass[conference,compsoc]{IEEEtran}
\IEEEoverridecommandlockouts

\usepackage{cite}
\usepackage{amsmath,amssymb,amsfonts}
\usepackage{algorithmic}
\usepackage{graphicx}
\usepackage{textcomp}
\usepackage{xcolor}
\usepackage{booktabs}
\usepackage{listings}

\newcommand{\rev}[1]{{\color{black}#1}}
\usepackage{needspace}

\def\BibTeX{{\rm B\kern-.05em{\sc i\kern-.025em b}\kern-.08em
    T\kern-.1667em\lower.7ex\hbox{E}\kern-.125emX}}
\begin{document}


\title{\rev{TopoIntent: Compiling Security Intent into Executable, Compliance-Checked Network Topologies}}


\author{
\IEEEauthorblockN{
Xiaokang Qu\IEEEauthorrefmark{1},
Jianliang Ma\IEEEauthorrefmark{2},
Zao Fan\IEEEauthorrefmark{2},
Tianshu Chu\IEEEauthorrefmark{2},
Tianlong Fan\IEEEauthorrefmark{1},
Linyuan Lü\IEEEauthorrefmark{1}
}

\IEEEauthorblockA{
\IEEEauthorrefmark{1}
School of Cyber Science and Technology, University of Science and Technology of China\\
Hefei, China\\
xkqu@mail.ustc.edu.cn, tianlong.fan@ustc.edu.cn, linyuan.lv@ustc.edu.cn
}

\IEEEauthorblockA{
\IEEEauthorrefmark{2}
Topsec Technologies Group Inc.\\
Beijing, China\\
ma\_jianliang@topsec.com.cn, fan\_zao@topsec.com.cn, chu\_tianshu@topsec.com.cn
}
}

\maketitle

\begin{abstract}
\rev{Enterprise security topology design begins before device configuration. Architects must translate business intent, regulatory requirements, and risk assumptions into zones, boundary devices, inter-zone paths, and access-control policies. Existing NetOps automation tools mostly operate after this design has been fixed: they translate policies into configurations or interpret existing diagrams, but provide limited support for producing a structured security topology from an underspecified natural-language request.}

\rev{We describe \textbf{TopoIntent}, a system that compiles security intent into executable, compliance-checked network topologies. TopoIntent uses a schema contract to constrain topology generation, retrieves reference architectures from a curated template library through dense-vector search, and applies staged fusion to separate intent-template alignment from IG-level security completion. It then checks the generated topology against CIS Controls v8.1.2 safeguards whose evidence is visible at the topology layer, records cases requiring manual review, and repairs structural gaps through additive schema-preserving edits. The final topology is exported to Mininet scripts with kernel-level \texttt{iptables} ACLs, so that reachability and allow/deny behavior can be exercised rather than inspected only as a static diagram.}

\rev{We build an evaluation set from reference security architecture diagrams because no public benchmark exists for this requirement-to-topology task. The retrieval reference set contains 22 templates and 44 synthetic intents across five scenarios; the held-out evaluation set contains 7 templates and 14 intents from finance and government scenarios excluded from the retrieval index. We evaluate TopoIntent with structural coverage, topology-visible CIS satisfaction, Mininet policy tests, and diagnostic feedback from Stage~5 to Stage~4. On the held-out set, additive repair raises first-pass topology-visible CIS satisfaction from 0.78/0.78 (IG2/IG3) to 1.00 in fewer than 1.5 rounds on average. In the end-to-end ablation, one feedback round raises the post-ACL policy pass rate from 0.78 to 0.88.}

\end{abstract}

\begin{IEEEkeywords}
Intent Networking, Topology Generation, Compliance, RAG, Network Security
\end{IEEEkeywords}

\section{Introduction}

\rev{Enterprise network security design is still largely a manual engineering task. Before a network is deployed, security architects must turn organizational requirements---for example, regulatory exposure, business functions, and risk tolerance---into a topology with explicit zones, boundary devices, access-control paths, and monitoring points. The design also has to be checked against security frameworks such as CIS Controls~\cite{cis_controls_v81}. In practice, engineers draft diagrams, compare them with compliance checklists, identify missing controls, and revise the design over multiple iterations. This work is slow, expensive, and difficult to scale as enterprise networks become larger and compliance obligations become more specific.}

\rev{LLMs have recently been used for network management tasks, including translating natural-language intents into device configurations~\cite{wang2024netconfeval, lira2024large, wei2025inta, tu2025intent}, generating NETCONF/YANG-based configurations~\cite{hollosi2024generative}, and supporting network operations through multi-agent workflows~\cite{Wang2025IntentDrivenNM}. These systems mainly address the implementation stage: the architecture is assumed to exist, and the model helps produce or modify device-level artifacts. The earlier design stage remains less explored. Given only a business-level requirement, there is still no standard way to synthesize a complete security-zone topology, check its structural compliance obligations, and test whether the resulting connectivity and policy artifacts behave as intended.}

\rev{This distinction matters. Existing intent-driven networking tools work within a known architecture~\cite{wang2024netconfeval, lira2024large, wei2025inta}. Formal synthesis systems such as Propane~\cite{Beckett2016DontMT}, Propane/AT~\cite{Beckett2017NetworkCS}, and NetComplete~\cite{ElHassany2018NetCompletePN} take formal specifications as input and synthesize routing configurations, not security architectures. GeNet~\cite{Ifland2024GeNetAM} accepts topology diagrams and assists engineers in modifying them, but it does not start from a free-form requirement and does not include compliance checking. We study a different problem: how to generate a security topology from natural language, preserve the user intent, check the topology against structural compliance constraints, and validate the generated reachability and access-control behavior in an executable environment.}

\rev{The difficulty comes from two sources. First, natural-language requirements are often incomplete. A request may specify a sector and a few critical services, but omit the DMZ, management plane, monitoring points, or boundary enforcement devices that a secure design would normally require. Second, the necessary design knowledge is distributed across reference architectures and compliance documents. A useful system must therefore combine intent understanding, design priors, compliance constraints, and executable validation without reducing the task to a single unconstrained LLM generation step.}

\rev{We present \textbf{TopoIntent}, an LLM-assisted system for intent-based security topology generation and compliance checking. TopoIntent follows a compiler-like workflow: it parses user intent into a schema-governed topology contract, retrieves relevant reference architectures from a curated template library, fuses retrieved templates with the user intent, checks the result against topology-visible CIS Controls v8.1.2 requirements, repairs structural gaps through diagnostic feedback, and exports the topology to Mininet~\cite{lantz2010network} for reachability and ACL validation. The output is not only a diagram-like topology, but a set of inspectable artifacts: JSON/XML topology records, generated policies, compliance decisions, repair history, visualization, and executable tests.}

We make the following contributions:

\begin{itemize}

\item \rev{\textbf{A security-design compilation pipeline.} TopoIntent maps natural-language requirements into a schema-constrained topology representation that can be retrieved against, fused, checked, repaired, and emulated through a single machine-readable contract.}

\item \rev{\textbf{Template-grounded topology synthesis.} We build a curated multi-industry template library and use dense retrieval to ground topology generation in reference security architectures. The ablation shows that retrieval-augmented fusion improves structural coverage over direct generation and over using retrieved templates without fusion.}

\item \rev{\textbf{CIS-guided structural checking and repair.} We integrate CIS Controls v8.1.2 as a scoped set of topology-visible safeguards, distinguish structural evidence from management-layer evidence, and use additive repair to address unsatisfied or manual-review findings while preserving the original topology elements.}

\item \rev{\textbf{Executable validation and diagnostic feedback.} We construct a held-out benchmark for the new task and evaluate the system with structural metrics, scoped CIS satisfaction, and Mininet/\texttt{iptables} tests. Stage~5 reports separate topology and ACL failure categories and feeds repairable diagnostics back to Stage~4.}

\end{itemize}

\rev{The evaluation reports both stage-level behavior and end-to-end ablations. A key point is that Stage~5 does not merely report whether tests pass; it separates missing paths from ACL-enforcement errors, which allows Stage~4 to repair the relevant part of the generated design rather than blindly changing the whole topology.}

The remainder of this paper is organized as follows.
Section~\ref{sec:preliminaries} introduces background concepts and
preliminaries.
Section~\ref{sec:overview} presents the system overview.
Section~\ref{sec:design} details the design of each TopoIntent module.
Section~\ref{sec:eval} describes our dataset construction, evaluation
methodology, and experimental results.
Section~\ref{sec:discussion} discusses limitations and threats to validity.
Section~\ref{sec:related} reviews related work.
Section~\ref{sec:conclusion} concludes the paper.

\section{Preliminaries}\label{sec:preliminaries}

\subsection{Security Zone-Based Network Architecture}

A security zone (also referred to as a security domain or network segment) is a logical partition of a network in which devices and services share a common security policy, trust level, and access control posture~\cite{Stouffer2023GuideTO}. Zone-based network design is the foundational approach recommended by
major security frameworks for structuring enterprise and institutional networks: rather than applying flat, perimeter-only defenses, the network is partitioned into multiple zones---such as an internet-facing
demilitarized zone (DMZ), an internal office zone, a data center zone, and a management zone---each separated by boundary enforcement devices such as firewalls and intrusion prevention systems (IPS).

A \emph{security topology} specifies (i) the set of security zones and their logical boundaries, (ii) the types and placement of security devices at zone boundaries, (iii) the inter-zone access control policies governing which traffic is permitted or denied, and (iv) the nodes (servers, endpoints, security appliances) assigned to each zone. Designing a compliant security topology requires translating organizational requirements and regulatory mandates into all four of these structural dimensions simultaneously---a task that demands both deep domain expertise and systematic compliance knowledge.

\subsection{CIS Controls v8.1.2 and Implementation Groups}

The CIS Controls~\cite{cis_controls_v81} (Center for Internet Security Controls, version 8.1.2) are a prioritized set of 18 security control domains and 153 specific safeguards that provide actionable guidance
for defending against the most prevalent cyber threats. Each safeguard is assigned to one of three \emph{Implementation Groups} (IGs) based on the risk profile and resource capacity of the
organization:

\begin{itemize}
    \item \textbf{IG1} covers the 56 foundational safeguards essential for every organization, regardless of size or sector. These address basic cyber hygiene such as asset inventory, access control, and data recovery.

    \item \textbf{IG2} extends IG1 with 74 additional safeguards targeting organizations that manage sensitive data or support critical services. Controls in this group address network monitoring, incident response, and security configuration management.

    \item \textbf{IG3} encompasses all 153 safeguards and is designed for organizations facing sophisticated, targeted attacks. It adds advanced capabilities such as penetration testing, application layer filtering, and security awareness training programs.
\end{itemize}

\rev{Only part of CIS can be judged from a topology. We therefore define a safeguard as \emph{topology-visible} when its evidence can be expressed through zones, boundary devices, remote-access paths, monitoring placement, exposed assets, or inter-zone filtering. This screening yields 22 safeguards, mainly from controls related to network infrastructure management, monitoring, access control, and traffic filtering. Safeguards that require asset inventories, operational procedures, vulnerability records, user training, or runtime configuration audits are outside the automated topology layer. Accordingly, TopoIntent does not claim organizational CIS certification; it checks the scoped set of CIS-derived structural requirements that can be represented in a security topology.}

\subsection{Retrieval-Augmented Generation}

Retrieval-Augmented Generation (RAG)~\cite{Lewis2020RetrievalAugmentedGF} augments the inference of a large language model by dynamically retrieving relevant documents from an external knowledge base and injecting
them into the model's context at query time. This approach addresses two fundamental limitations of parametric LLMs: knowledge staleness and hallucination. A comprehensive survey of RAG variants and applications is provided by Gao et al.~\cite{gao2023retrieval}.

Standard RAG systems encode a document corpus into dense vector representations using a text embedding model, index the vectors in a dedicated vector database, and at inference time retrieve the top-$k$ most semantically similar documents to the input query via approximate nearest-neighbor search. The retrieved documents are concatenated with the user query as context before being passed to the generative model.

\rev{TopoIntent uses retrieval in two places. In Stage~2, each reference template is encoded as a single vector over its concatenated fields and retrieved by cosine similarity; Section~\ref{subsec:stage-results} shows that this simple strategy performs better than the tested multi-field variants on the current template library. In Stage~4, CIS safeguards are not retrieved semantically. The scoped structural safeguards are stored in a control database and selected by IG level, which avoids omitting a required check because of embedding similarity.}

\section{System Overview}\label{sec:overview}

\subsection{Architecture Overview}

\begin{figure*}[t]
  \centerline{\includegraphics[width=\textwidth]{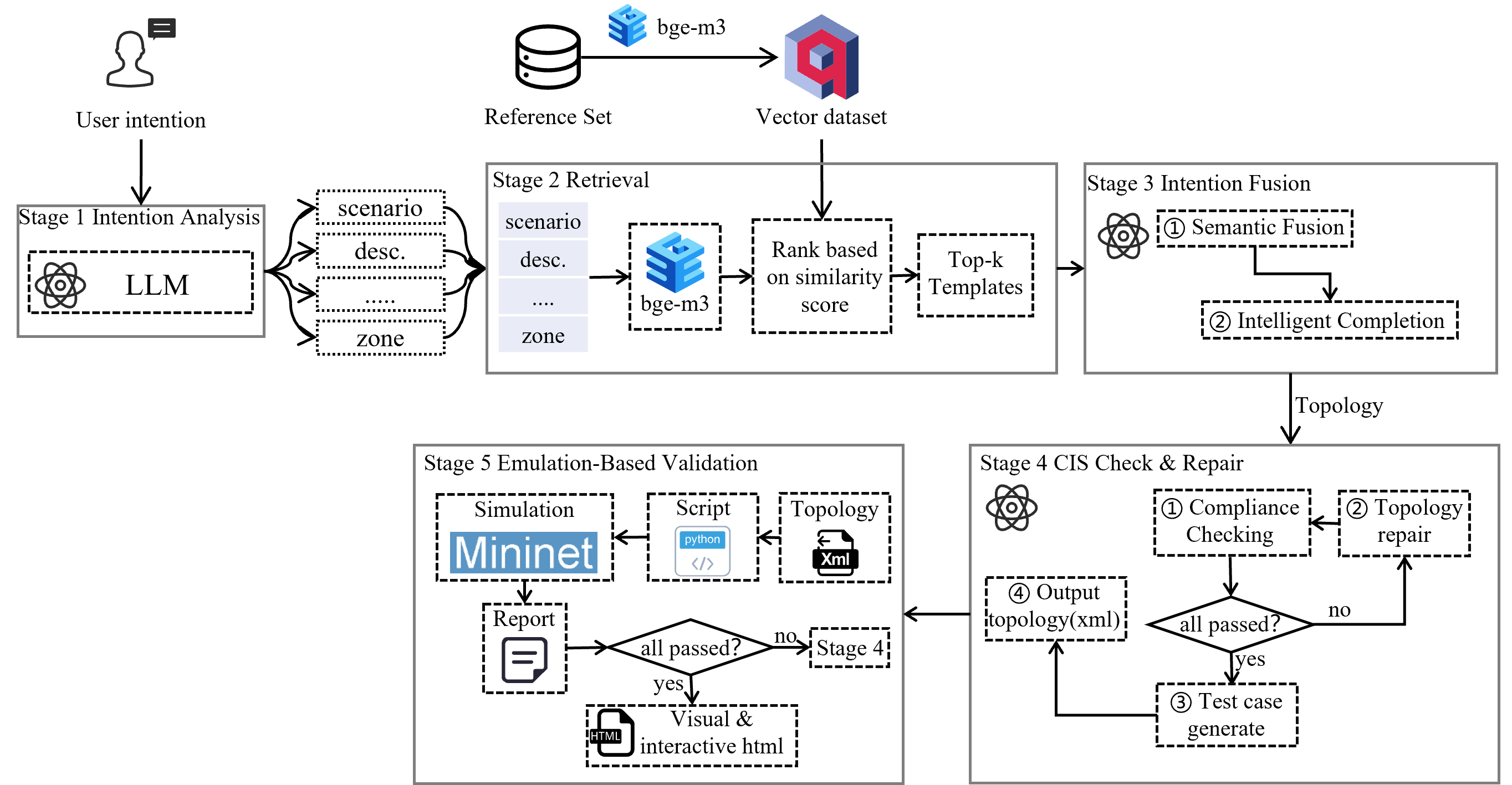}}
    \caption{System architecture of TopoIntent. The pipeline comprises five stages: intent analysis, template retrieval, intent fusion, CIS-aware compliance check and repair, and emulation-based validation.}
  \label{fig:overview}
\end{figure*}

\rev{TopoIntent transforms free-form security requirements into structured, compliance-checked security zone topologies.} As illustrated in Fig.~\ref{fig:overview}, the system comprises five sequential stages: Intent Analysis, Template Retrieval,
Intent Fusion, CIS-Aware Compliance Check \& Repair, and Emulation-Based Validation.
All inter-stage data exchange is governed by a unified structured data contract (\textsc{SchemaContract}), which defines the canonical JSON representation of a security topology and is described in Section~\ref{sec:design}.

\textbf{Stage 1 --- Intent Analysis.}
The user's natural language input is processed by an LLM-based intent parser that extracts an initial topology template conforming to \textsc{SchemaContract}. The parser normalizes scenario, zone, node, and link fields into the schema enumerations and produces a machine-readable artifact for retrieval and fusion.

\textbf{Stage 2 --- Template Retrieval.}
The structured intent JSON is used to query the TopoIntent Template Library, a curated collection of reference security topologies sourced from authoritative industry publications and vendor reference architectures.
Each template is encoded as a single dense vector over its concatenated fields using BGE-M3~\cite{Chen2024M3EmbeddingMM} and indexed in Qdrant. At query time, the Stage~1 output is encoded using the same
procedure and the top-$k$ most similar templates are retrieved via cosine similarity.

\textbf{Stage 3 --- Intent Fusion.}
The user intent JSON and the top-$k$ retrieved templates (default $k$ = 3) are fused through a two-stage LLM pipeline: (1)~\textit{Semantic Fusion}, which aligns and merges the two JSON structures without inference, preserving all user-specified fields verbatim; and (2)~\textit{Intelligent Completion}, which fills in missing fields
in accordance with the user's target CIS Controls Implementation Group (IG) level. If no IG level is specified, the stage produces two independent topology instances calibrated to IG2 and IG3 assurance levels
respectively. The output is a structurally complete topology JSON ready for compliance verification.

\textbf{Stage 4 --- CIS-Aware Compliance Check \& Repair.}
\rev{The generated topology is checked against the scoped CIS Controls v8.1.2 safeguards. The checker evaluates the topology JSON and labels each safeguard as \textit{satisfied}, \textit{unsatisfied}, or \textit{manual\_review}, using topology evidence rather than general product assumptions. Unsatisfied or uncertain safeguards are sent to an additive repair step that proposes the smallest schema-valid changes needed to provide the missing structural evidence. After repair, the system generates connectivity tests covering permitted flows, blocked flows across security boundaries, and paths through boundary enforcement devices.}
The repaired topology is exported as a structured XML artifact and rendered as an interactive HTML visualization.

\textbf{Stage 5 --- Emulation-Based Validation.}
The XML topology is converted into a Mininet emulation script that instantiates topology nodes as Linux network namespaces and realizes only node-level links as point-to-point links. Security zones remain logical metadata rather than emulated devices. The script first checks raw node-link reachability, then executes policy tests before and after installing \texttt{iptables} ACL rules derived from the generated policy set. The resulting JSON report records reachability, latency, ACL verdicts, and diagnostic failure categories. If validation failures indicate missing topology paths or inconsistent ACL behavior, the report can be fed back to Stage~4 to trigger targeted topology repair, forming a closed-loop refinement cycle.

\subsection{Design Principles}

\rev{TopoIntent follows four design principles.}

\textbf{Determinism First.}
\rev{LLMs are not used for tasks that can be handled by rules. Node attribute inference, data-flow classification, schema validation, name normalization, and simulation-script generation are implemented as deterministic procedures so that identical inputs produce identical outputs.}

\textbf{Contract-Driven Modularity.}
\rev{All modules exchange data through \textsc{SchemaContract}, a JSON schema that defines field semantics, type constraints, and referential-integrity rules. A module may enrich the topology only through fields admitted by the contract, which keeps intermediate results inspectable and prevents silent schema drift.}

\textbf{Artifact Persistence.}
\rev{Each stage writes structured logs and artifacts, including topology XML, interactive HTML, simulation scripts, repair records, and test reports. These files make it possible to inspect how a generated topology changed from intent parsing to executable validation.}

\textbf{Graceful Degradation.}
\rev{Local failures are handled explicitly. Malformed JSON is normalized when possible; unparseable configuration lines are logged as \texttt{unknown\_lines}; nodes without usable IP information are skipped during script generation with a warning rather than causing the entire run to fail.}

\section{Design of TopoIntent}\label{sec:design}

\subsection{Data Contract Design}\label{subsec:contract}

All inter-stage data exchange in TopoIntent is governed by \textsc{SchemaContract} (v2.1), a formally defined JSON schema that serves as the canonical representation of a security topology throughout the pipeline.
The schema comprises three field groups. The \textit{identity} group captures the scenario category (7 types),
a free-text description, and a structured topology summary. The \textit{spatial} group defines zones (each with a \texttt{zone\_id} and one of 11 \texttt{zone\_type} categories) and nodes (each with a \texttt{node\_id}, a \texttt{zone\_id} reference, and one of 20 \texttt{normalized\_type} device categories).
The \textit{connectivity} group encodes zone-level and node-level directed links, each annotated with a \texttt{link\_type} (9 categories) and an optional label.

\rev{Three structural invariants are enforced: (i)~ID-based referential integrity between nodes, zones, and links; (ii)~enumerated type constraints on \texttt{zone\_type}, \texttt{normalized\_type}, and \texttt{link\_type} to prevent semantic drift; and (iii)~separation between the base topology and later artifacts such as policies, compliance verdicts, and test cases. This separation lets each stage be inspected independently instead of collapsing topology synthesis, compliance checking, and executable validation into one opaque output.}

\subsection{Intent Analysis}\label{subsec:intent}

The intent analysis stage transforms a free-form natural language description of network security requirements into an initial topology template conforming to \textsc{SchemaContract} v2.1. The extraction is performed by Qwen2.5-72B-Instruct~\cite{Yang2024Qwen25TR} in a zero-shot setting. The full \textsc{SchemaContract} definition is embedded in the system prompt as a structural constraint, and the model is instructed to emit a single valid JSON object with no additional commentary.

The model is expected to populate all three field groups of the schema: the \textit{identity} group (scenario category, description, topology summary), the \textit{spatial} group (zones with \texttt{zone\_type} assignments and nodes with \texttt{normalized\_type} and \texttt{zone\_id} references), and the
\textit{connectivity} group (node-level and zone-level links with \texttt{link\_type} annotations).

A post-processing validation step handles imperfect model outputs: JSON is extracted from markdown fences if present; missing required fields are populated with typed defaults; out-of-vocabulary \texttt{scenario}, \texttt{zone\_type}, and \texttt{normalized\_type} values are remapped to their closest valid enumeration entry or normalized to \texttt{other}; and \texttt{zone\_id} references in \texttt{nodes} and \texttt{links} that do not correspond to any declared zone are corrected by assignment to the first available zone. If the model output cannot be parsed after two attempts, the stage raises a structured error and logs the raw outputs for inspection, while returning a minimal valid template to allow downstream stages to proceed.

Stage~1 outputs are evaluated against ground-truth templates on three metrics: \textit{scenario accuracy} (exact match), \textit{zone type recall} (fraction of ground-truth \texttt{zone\_type} values covered by the predicted zone set), and \textit{node type recall} (fraction of ground-truth \texttt{normalized\_type} values covered by the predicted node set).

\subsection{Template Retrieval}\label{subsec:retrieval}

The Template Library stores reference security topologies indexed in Qdrant as dense vector representations. Each template is encoded as a single dense vector by concatenating five textual fields---\textit{title}, \textit{description}, \textit{scenario}, \textit{topology\_summary}, and \textit{security}---and encoding the result with BGE-M3~\cite{Chen2024M3EmbeddingMM} for approximate nearest-neighbor search~\cite{johnson2019billion}.
The \textit{security} field is constructed by concatenating the template's zone type sequence and the deduplicated set of node types, providing a compact structural fingerprint of the topology's security architecture.

At query time, the Stage~1 output template is encoded using the same concatenation procedure to produce a single query vector. The top-$k$ most similar templates are retrieved via cosine similarity and returned to the fusion stage.
We compare this single-vector strategy against two multi-field variants in Section~\ref{sec:eval}: a uniform-weight variant that encodes each field independently and aggregates scores with equal weights, and a weighted variant with tuned per-field weights. Empirical results on the reference set show that single-vector retrieval achieves the highest Top-1 accuracy and MRR among the three variants (Section~\ref{subsec:stage-results}), which we attribute to the relatively small template library size (22 templates) where a concatenated embedding captures sufficient cross-field signal without requiring field decomposition.
Retrieval quality is measured by Top-$k$ accuracy ($k \in \{1,3,5\}$) and Mean Reciprocal Rank~(MRR).

\subsection{Intent Fusion}\label{subsec:fusion}

The intent fusion stage combines the extracted user intent JSON with the top-ranked retrieved template through two sequential LLM calls.

\subsubsection{Stage 3.1: Semantic Fusion}
The first call receives both the user intent JSON and the template JSON as a joint input. Its objective is structural alignment without inference: fields present in the user intent take precedence and are preserved verbatim; fields absent from the user intent but present in the template are used to fill the corresponding positions in the output. The model is explicitly prohibited from introducing fields not present in either input or from making semantic inferences beyond direct structural merging. The output is a structurally complete but not yet fully specified topology JSON.

\subsubsection{Stage 3.2: Intelligent Completion}
The second call receives the Stage~3.1 output and performs IG-level-aware completion. If the user has specified a target Implementation Group level (IG1, IG2, or IG3), the model
completes the topology according to the safeguard requirements of the specified IG level while preserving the simplified \textsc{SchemaContract}. Completion focuses on adding missing zones, nodes, node links, and zone-level logical relations using the allowed schema fields; access-control policies, compliance verdicts, and executable test cases are generated later by Stage~4 as separate artifacts rather than embedded in the base topology schema. If no IG level is specified, the stage produces two independent topology instances: one completed for IG2 and one completed for IG3, providing the user with topologies calibrated to two distinct IG levels. Throughout completion, the model is constrained to maintain ID-based referential integrity across zones, nodes, and links.

\subsection{CIS-Aware Compliance Check and Repair}\label{subsec:compliance}

\subsubsection{IG-Level Parameter and Control Selection}
\rev{Stage~4 receives the Stage~3 topology and a target IG level. If the user does not specify an IG level, the stage runs separately for IG2 and IG3 and produces two checked topology instances. The control database first filters CIS Controls v8.1.2 to the 22 safeguards with topology-level evidence. It then selects the IG-applicable subset: IG2 includes the structural safeguards assigned to IG1 or IG2, and IG3 includes the full structural subset.}

\subsubsection{Compliance Verification}
\rev{The selected safeguards are submitted to Qwen2.5-72B-Instruct together with the topology JSON. Each safeguard record contains its identifier, title, IG level, topology evidence fields, and a short predicate-style checking instruction. The checker returns a JSON object in which every safeguard has a \texttt{status} field (\textit{satisfied}, \textit{unsatisfied}, or \textit{manual\_review}), a confidence score in $[0,1]$, and evidence text pointing to the topology elements used in the decision. A normalization step inserts missing safeguards as \textit{manual\_review}, removes malformed fields, and clamps confidence values. The checker is therefore used as a topology-evidence assessor, not as an oracle for full CIS certification.}

\subsubsection{Iterative Additive Repair}
Controls classified as \textit{unsatisfied} or \textit{manual\_review} are forwarded to the repair step. Unlike full-topology rewriting, the repair call is explicitly constrained to return only \emph{additive} topology changes: new zones, new nodes, and new links that should be inserted to satisfy the identified findings. The additions are validated and merged into the current topology without modifying any existing elements. Following each repair round, the topology is re-verified against the full required control set.
\rev{The loop runs for at most \texttt{max\_repair\_rounds} iterations (default: 2), stopping early when all findings are resolved. To limit over-generation, the repair prompt asks for the smallest addition set that addresses the current findings; the merger rejects duplicate IDs, dangling references, and schema-invalid elements. Each round also writes a diff record listing the safeguards addressed, the additions applied, and the before/after counts for zones, nodes, zone links, and node links.}

\subsubsection{Stage~5 Diagnostic Feedback}
If Stage~5 emulation reveals policy failures or unexecutable test cases, the validation report is fed back to Stage~4 via a dedicated diagnostic repair interface. The interface converts the four repair-actionable Stage~5 failure categories---topology failures, ACL failures, ACL conflicts, and node-link failures---into structured repair findings and submits them to the additive repair call; unresolvable (\textit{skipped}) test endpoints are surfaced for manual review rather than fed into automated repair, since they typically reflect intent inconsistencies that topology edits cannot resolve. A subsequent CIS re-verification pass is then executed to ensure that Stage~5-driven topology modifications do not introduce new compliance gaps.

\subsubsection{Test Case Generation}
After all repair rounds complete, connectivity test cases are generated from the finalized topology. The generator produces a structured set of node-to-node verification cases covering permitted flows across explicit zone and node links, blocked flows across security boundaries, and traffic paths traversing boundary enforcement devices (firewalls, VPN gateways, IDS/IPS, WAF). Each test case specifies \texttt{source\_zone}, \texttt{target\_zone}, \texttt{source\_node}, \texttt{target\_node}, \texttt{protocol}, \texttt{port}, expected verdict (\textit{allow} or \textit{block}), and a natural language rationale. A post-generation normalization step discards any test case whose source or target node identifier does not exist in the topology node list, ensuring all generated cases are executable by Stage~5.

\subsubsection{XML Export}
The finalized topology, compliance verification results, repair history, and test cases are serialized into a structured XML artifact. The XML root element carries the scenario category and IG level as attributes and contains sections for \texttt{Domains} (zone definitions), \texttt{Nodes} (node definitions with zone references), \texttt{Links} (zone- and node-level links), \texttt{Policies} (test cases as \texttt{Allow}/\texttt{Deny} elements with protocol and port attributes), \texttt{Compliance} (per-control verdicts with evidence, missing reason, and remediation text), and \texttt{RepairHistory} (per-round repair diffs). This artifact is the handoff document consumed by Stage~5 for emulation-based validation.

Stage~4 is evaluated on three metrics: \textit{compliance satisfaction rate} (fraction of required safeguards classified as satisfied, averaged across samples), \textit{average repair rounds}, and \textit{average test cases generated per topology}.

\subsection{Emulation-Based Validation}\label{subsec:emulation}

\subsubsection{XML Parsing and Name Normalization}
Stage~5 reads the XML artifact produced by Stage~4 and reconstructs the topology into a generator-friendly representation. Node identifiers are normalized to valid Linux interface names (maximum 9 usable characters, to accommodate the \texttt{-ethN} suffix appended by Mininet) using a two-step procedure: non-alphanumeric characters are stripped and the result is lowercased; names exceeding the length budget are truncated and disambiguated by a 4-character MD5 suffix; subsequent collisions among truncated names are resolved by appending an incrementing numeric suffix.

\subsubsection{Topology Instantiation}
The emulation script instantiates all nodes as \texttt{LinuxRouter} hosts---Mininet hosts with IP forwarding enabled---connected by direct links using the per-link point-to-point addresses. After the network starts, each node's loopback address is configured on its \texttt{lo} interface, IP forwarding and reverse-path filtering settings are applied, and shortest-path routes to all peer loopback addresses are installed using BFS over the node-link graph. This per-node routing setup enables multi-hop end-to-end reachability
without requiring a dedicated router node.

\subsubsection{ACL Deployment}
\rev{Policies are compiled into an ACL table and installed as \texttt{iptables} rules on the Mininet namespaces corresponding to the relevant endpoints or boundary nodes available in the topology. For each tested flow, Stage~5 installs protocol- and port-aware \texttt{ACCEPT} or \texttt{DROP} rules and then adds fallback \texttt{DROP} rules for the same endpoint pair. The goal is to test the executable allow/deny semantics of the generated policy table. It is not intended to emulate vendor-specific firewall behavior.}

\subsubsection{Test Execution}
Two categories of tests are executed. \textit{Node-link connectivity tests} verify that each point-to-point
link in the topology is operationally reachable by pinging the peer interface address; results are recorded with pass/fail verdict and round-trip latency. \textit{Policy tests} execute each test case from the
\texttt{Policies} section twice: once before ACL deployment to test whether the underlying topology path is executable, and once after ACL deployment to test whether the generated ACL table enforces the expected verdict. For TCP test cases with a specified port, reachability is probed using a short-lived Python socket server on the destination host and a connection attempt from the source; the server accepts one connection and exits. For all other test cases, three ICMP ping packets are issued from the source to the destination loopback address. In both cases, the observed outcome (\textit{allow} or \textit{deny}) is compared against the expected verdict from the test case, and a pass/fail judgment is recorded along with latency. Test cases whose source or destination cannot be resolved to an instantiated host are marked as \textit{skipped} and counted separately in the summary.

\subsubsection{Report Generation}
All results are serialized to a structured JSON report containing aggregate summaries, per-link reachability and latency records, pre-ACL and post-ACL policy-test records, and a diagnostic section. The diagnostic section explicitly separates topology failures, ACL failures, ACL conflicts, node-link failures, and unresolvable (\textit{skipped}) test endpoints, allowing Stage~4 to repair structural connectivity issues without mistaking them for policy generation errors.

\section{Evaluation}\label{sec:eval}

\subsection{Dataset Construction}\label{subsec:dataset}
\begin{figure*}[t]
  \centerline{\includegraphics[width=\textwidth]{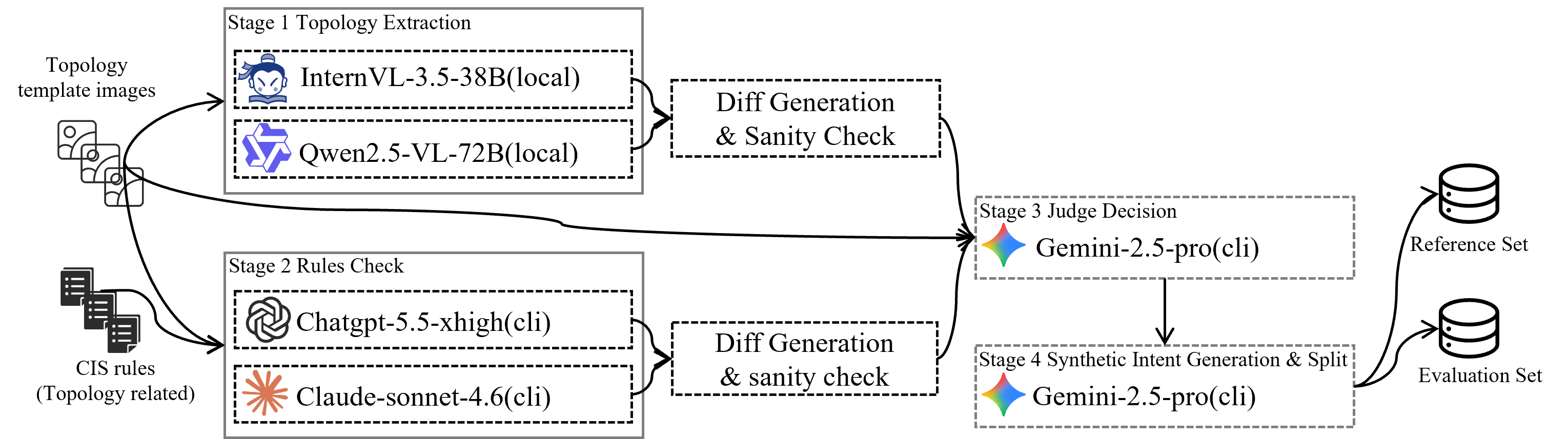}}
  \caption{\rev{Dataset construction pipeline. Stage~D1 extracts topology structures from reference architecture images using two VLMs in parallel; Stage~D2 annotates topology-visible CIS Controls evidence using two LLMs; Stage~D1 and Stage~D2 outputs are reconciled via diff generation and sanity check; Stage~D3 uses Gemini-2.5-Pro as a judge to produce the final ground-truth template; Stage~D4 generates synthetic user intents and splits the dataset into reference and held-out evaluation sets.}}
  \label{fig:dataset}
\end{figure*}

Constructing a high-quality ground-truth topology dataset from authoritative reference architecture diagrams requires resolving two independent challenges: accurate structural extraction from visual diagrams, and reliable CIS Controls compliance annotation. Both tasks are subjective and error-prone when performed by a single model, so we adopt a multi-model consensus pipeline with a judge-arbitration step, as illustrated in Fig.~\ref{fig:dataset}.

\subsubsection{Stage D1: Dual-Model Topology Extraction}
Each reference architecture diagram is independently processed by two vision-language models (VLMs): InternVL3.5-38B~\cite{Wang2025InternVL35AO} and Qwen2.5-VL-72B~\cite{qwen2.5-VL}, both served locally via vLLM. Each model is prompted to extract only the visually observable topology structure---zones, nodes, and links---into a \textsc{SchemaContract}-conforming JSON, without inferring hidden devices, IP addresses, or implicit security functions. A conservative extraction protocol is enforced: grouped nodes represent repeated identical devices; zone-level connections are captured as \texttt{zone\_links} rather than forced into
\texttt{node\_links}; and all \texttt{zone\_type}, \texttt{normalized\_type}, and \texttt{link\_type} values are constrained to the predefined schema enumerations. Each output is validated against referential integrity rules before being accepted.

\subsubsection{Stage D2: Dual-Model CIS Compliance Annotation}
In parallel, each reference architecture diagram is independently evaluated by two large language models via CLI: GPT-5.5 (reasoning effort = xhigh) and Claude Sonnet~4.6. Each model assesses which topology-visible CIS Controls v8.1.2 safeguards are satisfied, rejected, or require manual review, following conservative matching rules that prohibit inferring compliance from product names alone and require explicit visual evidence for each safeguard.

\subsubsection{Diff Generation and Sanity Check}
Before judge arbitration, pairwise diffs are computed between the two Stage~D1 topology outputs and between the two Stage~D2 CIS outputs. The topology diff computes signature-based agreement and disagreement on zones, nodes, and links, and flags schema violations in either output. The CIS diff classifies each safeguard into consensus-matched, consensus-rejected, hard-conflict (one matched, one rejected), or soft-conflict (involving manual review) categories, and surfaces safeguards with large confidence divergence. These diffs serve as structured focus documents for the judge.

\subsubsection{Stage D3: Judge Arbitration}
Gemini~2.5~Pro~\cite{comanici2025gemini} acts as the final judge, receiving the original topology image, both Stage~D1 topology candidates, both Stage~D2 CIS candidates, and the precomputed diffs as joint input. The judge is instructed to use the image as the primary source of truth, resolve conflicts by visible evidence rather than product capability assumptions, and produce a single final \textsc{SchemaContract}-conforming topology template and a complete CIS compliance decision in which every safeguard from Stage~D2 is classified exactly once into matched, rejected, or manual-review. Outputs are validated against full referential integrity and schema enum constraints before acceptance; failed outputs trigger up to two retries.

\subsubsection{Stage D4: Synthetic Intent Generation and Dataset Split}
For each finalized ground-truth topology, Gemini~2.5~Pro generates two synthetic user intent descriptions per template---one business-oriented and one architecture-oriented---by conditioning on the topology image, the final topology template, and the CIS compliance result. Generation constraints prohibit leaking internal schema identifiers, CIS safeguard IDs, or template IDs into the generated text, and enforce a word count range of 25--110 words per intent.

\rev{The finalized templates are split at the template level into a \textit{reference set} and a \textit{held-out evaluation set}. The reference set is indexed by the Template Library and used by Stage~2. The evaluation set is never indexed. We assign the finance and government scenarios to the holdout split because they carry stronger compliance expectations and make retrieval leakage especially problematic. This choice is stricter than a random split: during evaluation, the system must adapt design patterns from other sectors instead of retrieving near-duplicates from the same scenario family.}

\subsection{Experimental Setup}\label{subsec:experimental-setup}

\subsubsection{Hardware and Runtime Environment}
\rev{All experiments are run on a local server with 8$\times$ NVIDIA RTX~4090 GPUs. LLMs and embedding models are served through vLLM~\cite{kwon2023efficient}. Mininet scripts are executed on the same host with \texttt{iptables} enabled.}

\subsubsection{Model and System Configuration}
Qwen2.5-72B-Instruct~\cite{Yang2024Qwen25TR} serves as the primary instruction model across all LLM-driven stages: intent analysis, intent fusion, CIS verification, repair, and test-case generation.
BGE-M3~\cite{Chen2024M3EmbeddingMM} is used as the dense encoder for single-vector template retrieval, and Qdrant is used as the vector database for the Template Library. The \textsc{SchemaContract} and all post-processing validators are
shared across stages, ensuring that intermediate artifacts remain machine-checkable throughout the pipeline.

\subsubsection{Baselines}
\rev{We compare the full system with four ablations, each removing one part of the pipeline.}
\textbf{Direct}: the user's natural language input is passed directly to the LLM with the \textsc{SchemaContract} definition as the only constraint; no intent analysis, retrieval, fusion, compliance verification, or emulation is performed. \textbf{RAG w/o Fusion}: Stage~1 intent analysis and Stage~2 template retrieval are performed, but the retrieved template is used directly as output without Stage~3 intent-template merging. \textbf{RAG w/o Repair}: the full Stage~1--3 pipeline is executed, followed by Stage~4 CIS verification; however, the iterative repair loop is skipped and the first-pass topology is exported directly without compliance-driven correction. \textbf{RAG w/o Feedback}: the complete Stage~1--4 pipeline is executed including iterative repair, but Stage~5 emulation results are not fed back to Stage~4 for further repair. \textbf{Full TopoIntent}: all five stages are executed, including Stage~5 diagnostic feedback to Stage~4.

\subsection{Evaluation Metrics}\label{subsec:metrics}

We define metrics at four evaluation points along the pipeline.

\textbf{Stage~1 intent analysis} is evaluated on the reference set with scenario accuracy (exact match against ground-truth scenario label), zone-type recall (fraction of ground-truth \texttt{zone\_type} values present in the generated topology), and node-type recall (fraction of ground-truth \texttt{normalized\_type} values covered).

\textbf{Stage~2 retrieval} is evaluated on the reference set using Top-$k$ accuracy ($k \in \{1,3,5\}$) and mean reciprocal rank (MRR). Since evaluation templates are excluded from the retrieval database, exact-template Top-$k$ accuracy is not reported on the evaluation set.

\rev{\textbf{Stage~4 compliance} is evaluated on both IG2 and IG3 runs using topology-visible CIS satisfaction rate before and after the repair loop, average repair rounds, and average number of generated test cases.}

\rev{\textbf{Stage~5 executable validation} is evaluated on initial and post-feedback runs for both IG2 and IG3 using node-link reachability, pre-ACL path-executability pass rate, post-ACL policy pass rate, average latency, and count of diagnostic failures.}

\rev{Security topology generation is not a one-to-one reconstruction task: several designs may satisfy the same intent. We therefore use recall-oriented structural metrics to check whether required zone and device classes are covered, and rely on scoped compliance and executable validation metrics to test whether the generated structure supports the intended security behavior. Repair-history diffs report any additive expansion introduced by the compliance loop.}

\subsection{Stage-Level Results}\label{subsec:stage-results}

Table~\ref{tab:stage1} reports intent analysis results on the reference set. \textit{Scenario Accuracy} measures exact match between the predicted scenario label and the ground-truth label. \textit{Zone Type Recall} measures the fraction of ground-truth \texttt{zone\_type} values that appear in the predicted topology's zone set, reflecting how well the model captures the required security zone categories. \textit{Node Type Recall} measures the fraction of ground-truth \texttt{normalized\_type} values covered by the predicted node set, reflecting device-class coverage relative to the reference design. The high scenario accuracy (0.98) confirms that the model reliably identifies the deployment context; the lower node type recall (0.53) is expected, as natural language descriptions typically do not enumerate all device classes explicitly.

\begin{table}[t]
\caption{Stage~1 Intent Analysis Results (Reference Set, $n=44$)}
\label{tab:stage1}
\centering
\small
\begin{tabular}{lc}
\toprule
\textbf{Metric} & \textbf{Value} \\
\midrule
Scenario Accuracy  & 0.98 \\
Zone Type Recall   & 0.68 \\
Node Type Recall   & 0.53 \\
\bottomrule
\end{tabular}
\end{table}

\begin{table}[t]
\caption{Stage~2 Retrieval Ablation (Reference Set, $n=44$)}
\label{tab:stage2}
\centering
\small
\begin{tabular}{lcccc}
\toprule
\textbf{Retrieval Strategy} & \textbf{Top-1} & \textbf{Top-3} 
  & \textbf{Top-5} & \textbf{MRR} \\
\midrule
Multi-field uniform  & 0.60 & 0.93 & 1.0 & 0.75 \\
Multi-field weighted & 0.57 & 0.93 & 0.98 & 0.75 \\
Single-vector        & 0.75 & 0.91 & 1.0 & 0.84 \\
\bottomrule
\end{tabular}
\end{table}

Table~\ref{tab:stage2} compares three retrieval strategies on the reference set. \textit{Top-$k$ accuracy} measures whether the ground-truth template appears within the top-$k$ retrieved results; \textit{MRR} rewards strategies that rank the ground-truth template higher. Single-vector retrieval, which encodes each template as a single concatenated embedding, achieves the highest Top-1 accuracy (0.75) and MRR (0.84), outperforming both multi-field variants on the current 22-template library. We attribute this to the small library size, where a concatenated embedding captures sufficient cross-field signal without requiring field decomposition. All three strategies converge at Top-5, suggesting that the correct template is consistently retrievable within the top-5 candidates regardless of encoding strategy.

\rev{Table~\ref{tab:stage4} reports scoped CIS results before and after iterative repair. \textit{CIS Sat. (Before Repair)} is the fraction of required structural safeguards classified as satisfied on the first pass; \textit{CIS Sat. (After Repair)} is the same fraction after repair. \textit{Avg. Repair Rounds} records the number of verify-repair iterations per sample, and \textit{Avg. Test Cases} records the number of executable connectivity tests generated from the final topology. The repair loop raises satisfaction from 78\% to 100\% for both IG2 and IG3 in fewer than 1.5 rounds on average. This result concerns the selected topology-visible safeguards only; executable behavior is evaluated separately in Stage~5.}

\begin{table}[htbp]
\caption{Stage~4 CIS Compliance Verification and Repair Results (Evaluation Set, $n=14$)}
\label{tab:stage4}
\centering
\small
\begin{tabular}{lcc} 
\toprule
\textbf{Metric} & \textbf{IG2} & \textbf{IG3} \\
\midrule
CIS Sat. (Before Repair) & 0.78  & 0.78  \\
CIS Sat. (After Repair)  & 1.00  & 1.00  \\
Avg. Repair Rounds       & 1.43  & 1.36  \\
Avg. Test Cases          & 41.14 & 41.14 \\
\bottomrule
\end{tabular}
\end{table}

\subsection{Executable Validation and Feedback}\label{subsec:stage5}

\rev{Table~\ref{tab:stage5} summarizes the Mininet validation results. \textit{Node-Link Reach.} measures whether declared \texttt{node\_links} become reachable point-to-point links. \textit{Path Exec. (Pre-ACL)} measures whether each policy test has an underlying path before any \texttt{iptables} rule is installed; in this phase every expected verdict is treated as \textit{allow}, so block-flow tests are used only to check path availability. \textit{Post-ACL Pass} measures whether the observed result after ACL installation matches the intended \textit{allow} or \textit{block} verdict. For this reason, post-ACL pass can exceed pre-ACL path executability: a blocked flow is a failure before policy enforcement but a success after the corresponding ACL is installed. The 100\% node-link reachability indicates that Stage~4 produces Mininet-instantiable node-level connectivity. Remaining post-ACL failures come from policy-table behavior rather than missing physical links, which is why Stage~5 diagnostics are useful for targeted repair.}

\begin{table*}[!h]
\caption{Stage~5 Executable Validation Results with One Feedback Round (Evaluation Set, $n{=}14$).}
\label{tab:stage5}
\centering
\small
\begin{tabular}{lccccc}
\toprule
\textbf{Setting} & \textbf{Node-Link Reach.}
                 & \rev{\textbf{Path Exec.}} & \textbf{Post-ACL Pass}
                 & \textbf{Avg. Latency (ms)} & \textbf{Diag. Failures}\\
\midrule
IG2 Initial        & 1.00 & 0.64 & 0.87 & 0.06 & 214\\
IG2 After Feedback & 1.00 & 0.74 & 0.86 & 0.07 & 165\\
IG3 Initial        & 1.00 & 0.67 & 0.87 & 0.06 & 189\\
IG3 After Feedback & 1.00 & 0.77 & 0.89 & 0.07 & 147\\
\bottomrule
\end{tabular}
\end{table*}

\subsection{End-to-End Ablation}\label{subsec:ablation}

Table~\ref{tab:ablation} reports end-to-end results across five system variants on the evaluation set. \textit{Zone Type Recall} and \textit{Node Type Recall} measure structural faithfulness to the ground-truth topology, using the same definitions as in Stage~1 evaluation. \textit{CIS Sat. Rate} measures the fraction of required IG2 and IG3 safeguards satisfied in the final output topology, averaged across both IG levels; `--' indicates that the variant does not include CIS verification. \textit{Post-ACL Pass Rate} measures the fraction of generated policy test cases that enforce the expected verdict in Mininet after ACL deployment; `--' indicates that the variant does not include Stage~5 emulation. 
Each variant incrementally adds one pipeline component, allowing the contribution of each stage to be isolated: Direct establishes the LLM generation baseline without any structured pipeline; RAG w/o Fusion adds template retrieval but skips intent-template merging; RAG w/o Repair adds the full fusion pipeline but skips compliance repair; RAG w/o Feedback adds repair but disables the Stage~5 diagnostic loop; Full TopoIntent enables the complete pipeline. Notably, RAG w/o Fusion achieves lower zone-type recall than Direct (0.54 vs.\ 0.63), which we attribute to the scenario mismatch between the reference set and the evaluation set: retrieved templates are drawn from five non-overlapping scenarios, and using them directly without fusion propagates structural patterns misaligned with the finance and government evaluation topologies.
\rev{The ablation clarifies the role of each component. Fusion improves structural recall over directly using retrieved templates. Repair raises scoped CIS satisfaction from 0.70 to 1.00. Adding Stage~5 feedback then increases the post-ACL pass rate from 0.78 to 0.88, showing that executable diagnostics catch policy-level problems that the structural checker does not see. This ablation-level gain should not be confused with the per-IG before/after rows in Table~\ref{tab:stage5}, where feedback mainly reduces diagnostic failures and changes the aggregate post-ACL rate only slightly. Node-type recall is the only non-monotonic metric (0.82 for RAG w/o Feedback and 0.77 for Full TopoIntent), likely because a feedback repair can substitute a structurally equivalent device class while preserving the policy behavior measured by Stage~5.}

\begin{table*}[t]
 \caption{End-to-End Ablation Results (Evaluation Set, $n{=}14$).
  A check mark indicates the component is enabled in that variant.
  \textbf{Pipeline components.} \textbf{S1}/\textbf{S2}/\textbf{S3} are
  Stages 1--3 of the pipeline; \textbf{Verify} and \textbf{Repair} are
  Stage~4 CIS verification and the iterative additive-repair loop;
  \textbf{Emul} is Stage~5 executable emulation; \textbf{FB} is the
  Stage~5$\to$Stage~4 diagnostic feedback loop.
  \textbf{Metrics.} \textbf{Zone} and \textbf{Node} are zone-type and
  node-type recall against the ground-truth topology; \textbf{CIS} is
  the CIS satisfaction rate (fraction of required IG2/IG3 safeguards
  satisfied, averaged across both IG levels); \textbf{Post-ACL} is the
  \rev{post-ACL policy pass rate measured in Mininet emulation.}
  A dash (--) marks metrics not applicable to a variant.}
\label{tab:ablation}
\centering
\small
\setlength{\tabcolsep}{3.5pt}
\begin{tabular}{l ccccccc cccc}
\toprule
 & \multicolumn{7}{c}{\textbf{Pipeline Components}}
 & \multicolumn{4}{c}{\textbf{Metrics}} \\
\cmidrule(lr){2-8} \cmidrule(lr){9-12}
\textbf{Variant}
 & \textbf{S1} & \textbf{S2} & \textbf{S3}
 & \textbf{Verify} & \textbf{Repair} & \textbf{Emul} & \textbf{FB}
 & \textbf{Zone} & \textbf{Node} & \textbf{CIS} & \textbf{Post-ACL} \\
\midrule
Direct           &            &            &            &            &            &            &            & 0.63 & 0.61 & --   & --  \\ \addlinespace[3pt]
RAG w/o Fusion   & \checkmark & \checkmark &            &            &            &            &            & 0.54 & 0.63 & --   & --  \\ \addlinespace[3pt]
RAG w/o Repair   & \checkmark & \checkmark & \checkmark & \checkmark &            &            &            & 0.75 & 0.73 & 0.70 & --  \\ \addlinespace[3pt]
RAG w/o Feedback & \checkmark & \checkmark & \checkmark & \checkmark & \checkmark & \checkmark &            & 0.74 & 0.82 & 1.00 & 0.78 \\ \addlinespace[3pt]
Full TopoIntent  & \checkmark & \checkmark & \checkmark & \checkmark & \checkmark & \checkmark & \checkmark & 0.76 & 0.77 & 1.00 & 0.88 \\
\bottomrule
\end{tabular}
\end{table*}

\subsection{Qualitative Case Study}\label{subsec:case-study}

We examine one held-out sample, \texttt{Financial-4\_req1}, to illustrate TopoIntent's generation process. This sample was excluded from the reference template library. The user intent requires a new branch office with high availability, voice service continuity (PSTN failover on primary link failure), SIP-based voice, centralized authentication, and security logging.

\begin{figure*}[t]
  \centerline{\includegraphics[width=\textwidth]{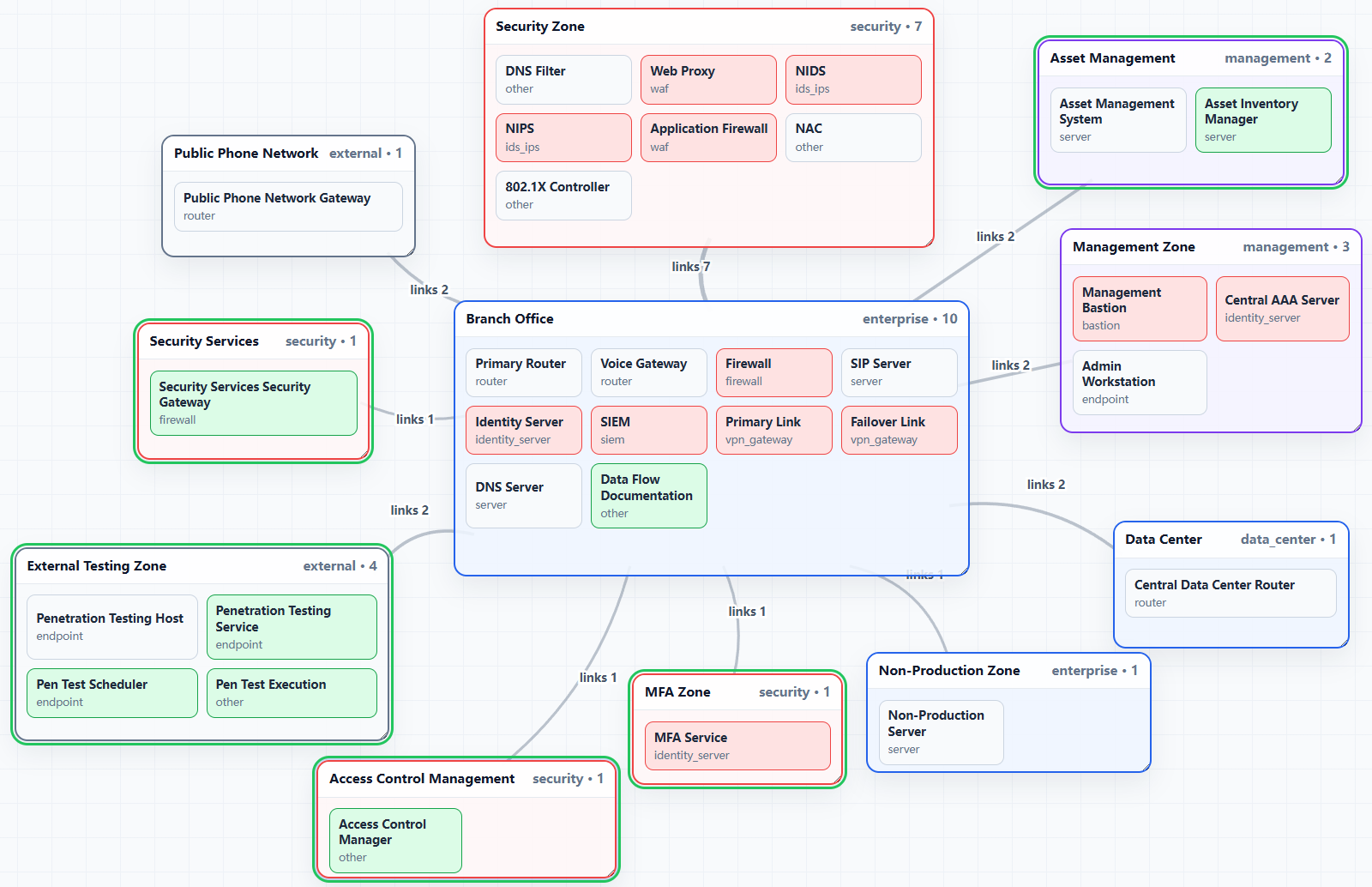}}
    \caption{Case study visualization for \texttt{Financial-4\_req1}. Green regions indicate advanced compliance and management zones introduced in IG3.}
  \label{fig:case-study}
\end{figure*}

As shown in Figure~\ref{fig:case-study}, TopoIntent produces an IG3 topology with clear zone-based segmentation, expanding from 6 zones (25 nodes) in the baseline IG2 design to 11 zones (32 nodes). The expansion introduces dedicated Security, Management, MFA, Asset Management, and External Testing zones.

The diagnostic feedback loop proves effective: the initial IG2 topology achieved a post-ACL pass rate of 0.72. After one feedback round, the pass rate reached 1.00. For IG3, the pass rate improved from 0.95 to 1.00. Figure~\ref{fig:case-study} visualizes the final topology, with green regions highlighting newly introduced compliance and management zones.

\section{Discussion and Limitations}\label{sec:discussion}

\rev{TopoIntent shows that natural-language security requirements can be turned into schema-governed topologies, checked against a scoped set of CIS-derived structural safeguards, and exercised in an executable network emulator. The contribution is the decomposition of the design task into inspectable steps: contract-based topology synthesis, template grounding, structural repair, and policy validation. Several boundaries of the current system should be kept in view.}

\rev{\textbf{Scope of compliance claims.}
CIS Controls v8.1.2 is used as a normative source, but TopoIntent does not certify full organizational CIS compliance. Controls depending on inventories, procedures, vulnerability-management records, training, or live configuration audits are outside the topology layer. The reported CIS rates therefore refer only to the selected safeguards whose evidence can be represented in the generated topology.}

\rev{\textbf{LLM judgments.}
The pipeline constrains LLM outputs with schemas, retries, normalization, deterministic post-processing, multi-model dataset construction, and persistent evidence records. Still, the checker and repair stages can make judgment errors, especially when a safeguard is only weakly implied by a topology. A natural extension is to replace more of these checks with deterministic graph and policy predicates once the structural interpretation of each safeguard is fixed.}

\rev{\textbf{Emulation fidelity.}
Mininet validation gives executable evidence for paths and allow/deny policy behavior, but it is not a production deployment. The emulator represents nodes as Linux namespaces and policies as \texttt{iptables} rules. This is sufficient for checking consistency between the generated topology and the generated policy table, but it does not model stateful firewall semantics, cloud control planes, proprietary appliances, or application-layer inspection.}

\rev{\textbf{Dataset scale.}
The current dataset is intentionally held out at the scenario level to reduce retrieval leakage, but it remains modest. The experiments should be read as an auditable first benchmark for requirement-to-security-topology generation, not as full coverage of enterprise architectures. Larger evaluations should include more sectors, larger diagrams, real practitioner-written requirements, and independent expert review of the compliance labels.}

\rev{\textbf{Additive repair.}
Additive repair preserves provenance because user-specified topology elements are not silently deleted. The tradeoff is that the system may add redundant devices or links when the checker is conservative. TopoIntent limits this by requesting minimal additions, rejecting invalid elements, and recording diffs, but future versions should add explicit repair costs or graph-edit penalties.}

\rev{\textbf{Feedback convergence.}
Stage~5 feedback supplies useful diagnostics, but it does not guarantee that one round will make every policy test pass. Some failures come from contradictory test cases, underspecified intent, or design choices that require human judgment. In these cases the system should expose the conflict rather than force an unsafe automatic repair.}

\section{Conclusion}\label{sec:conclusion}

\rev{We presented TopoIntent, a system for compiling natural-language security intent into executable, compliance-checked network topologies. The system converts free-form requirements into a schema-governed topology contract, grounds generation with retrieved reference architectures, repairs structural gaps against a scoped set of CIS Controls v8.1.2 safeguards, and validates reachability and ACL behavior in Mininet/\texttt{iptables}. This design links static topology generation with executable evidence while keeping topology, compliance decisions, repairs, and policy tests as separate inspectable artifacts. On the held-out evaluation set, additive repair raises topology-visible CIS satisfaction from 0.78/0.78 (IG2/IG3) to 1.00, and the end-to-end ablation shows that one Stage~5 feedback round raises post-ACL pass rate from 0.78 to 0.88. The results suggest that requirement-to-security-topology generation can be treated as an end-to-end design workflow, while also making clear where structural checking ends and production-grade deployment validation must begin.}
\section{LLM Usage Statement}

\rev{Large language models were used in two ways. First, they are part of the TopoIntent system: Qwen2.5-72B-Instruct is used for intent analysis, topology fusion and completion, CIS-oriented checking and repair, test-case generation, and diagnostic-feedback-based refinement; BGE-M3 is used for embedding-based template retrieval. These uses are described in the methodology sections. Generated artifacts were checked through schema validation, scoped CIS verification, Mininet-based tests, and manual inspection of representative outputs.}

\rev{Second, LLMs were used as editorial aids during manuscript preparation, including wording and organization. The authors inspected the resulting text and take full responsibility for the correctness, originality, and integrity of all claims, tables, figures, experiments, and conclusions. The data used in this work consists of synthetic network-intent samples and topology templates constructed for this study; no private user data was used. We report model configuration, prompts, schema contracts, evaluation scripts, and generated artifacts to reduce reproducibility risks from model updates, decoding randomness, prompt sensitivity, and deployment differences.}

\bibliographystyle{IEEEtran}
\bibliography{reference}

@techreport{cis_controls_v81,
  author      = {{Center for Internet Security}},
  title       = {{CIS Critical Security Controls Version 8.1}},
  institution = {Center for Internet Security (CIS)},
  year        = {2024},
  type        = {Technical Report},
  url         = {https://www.cisecurity.org/controls/v8-1},
  note        = {Accessed: May 2026}
}

@article{wang2024netconfeval,
  title={Netconfeval: Can llms facilitate network configuration?},
  author={Wang, Changjie and Scazzariello, Mariano and Farshin, Alireza and Ferlin, Simone and Kosti{\'c}, Dejan and Chiesa, Marco},
  journal={Proceedings of the ACM on Networking},
  volume={2},
  number={CoNEXT2},
  pages={1--25},
  year={2024},
  publisher={ACM New York, NY, USA}
}

@article{lira2024large,
  title={Large language models for zero touch network configuration management},
  author={Lira, Oscar G and Caicedo, Oscar M and da Fonseca, Nelson LS},
  journal={IEEE Communications Magazine},
  volume={63},
  number={7},
  pages={146--153},
  year={2024},
  publisher={IEEE}
}

@inproceedings{wei2025inta,
  title={Inta: Intent-based translation for network configuration with llm agents},
  author={Wei, Yunze and Xie, Xiaohui and Hu, Tianshuo and Zuo, Yiwei and Chen, Xinyi and Chi, Kaiwen and Cui, Yong},
  booktitle={2025 IEEE 33rd International Conference on Network Protocols (ICNP)},
  pages={1--16},
  year={2025},
  organization={IEEE}
}

@article{tu2025intent,
  title={Intent-Based Network Configuration Using Large Language Models},
  author={Tu, Nguyen and Nam, Sukhyun and Hong, James Won-Ki},
  journal={International Journal of Network Management},
  volume={35},
  number={1},
  pages={e2313},
  year={2025},
  publisher={Wiley Online Library}
}

@inproceedings{hollosi2024generative,
  title={Generative ai for low-level netconf configuration in network management based on yang models},
  author={Holl{\'o}si, Gergely and Ficzere, D{\'a}niel and Varga, P{\'a}l},
  booktitle={2024 20th International Conference on Network and Service Management (CNSM)},
  pages={1--7},
  year={2024},
  organization={IEEE}
}

@article{Wang2025IntentDrivenNM,
  title={Intent-Driven Network Management with Multi-Agent LLMs: The Confucius Framework},
  author={Zhaodong Wang and Samuel Lin and Guanqing Yan and Soudeh Ghorbani and Minlan Yu and Jiawei Zhou and Nathan Hu and Lopa Baruah and Sam Peters and Srikanth Kamath and Je-Loon Yang and Ying Zhang},
  journal={Proceedings of the ACM SIGCOMM 2025 Conference},
  year={2025},
  url={https://api.semanticscholar.org/CorpusID:280953986}
}

@article{Beckett2016DontMT,
  title={Don't Mind the Gap: Bridging Network-wide Objectives and Device-level Configurations},
  author={Ryan Beckett and Ratul Mahajan and Todd D. Millstein and Jitendra Padhye and David Walker},
  journal={Proceedings of the 2016 ACM SIGCOMM Conference},
  year={2016},
  url={https://api.semanticscholar.org/CorpusID:263266303}
}

@article{Beckett2017NetworkCS,
  title={Network configuration synthesis with abstract topologies},
  author={Ryan Beckett and Ratul Mahajan and Todd D. Millstein and Jitendra Padhye and David Walker},
  journal={Proceedings of the 38th ACM SIGPLAN Conference on Programming Language Design and Implementation},
  year={2017},
  url={https://api.semanticscholar.org/CorpusID:26265051}
}

@inproceedings{ElHassany2018NetCompletePN,
  title={NetComplete: Practical Network-Wide Configuration Synthesis with Autocompletion},
  author={Ahmed El-Hassany and Petar Tsankov and Laurent Vanbever and Martin T. Vechev},
  booktitle={Symposium on Networked Systems Design and Implementation},
  year={2018},
  url={https://api.semanticscholar.org/CorpusID:4787777}
}

@article{Ifland2024GeNetAM,
  title={GeNet: A Multimodal LLM-Based Co-Pilot for Network Topology and Configuration},
  author={Beni Ifland and Elad Duani and Rubin Krief and Miro Ohana and Aviram Zilberman and Andres Murillo and Ofir Manor and Ortal Lavi and Hikichi Kenji and Asaf Shabtai and Yuval Elovici and Rami Puzis},
  journal={2025 IEEE 45th International Conference on Distributed Computing Systems Workshops (ICDCSW)},
  year={2024},
  pages={117-122},
  url={https://api.semanticscholar.org/CorpusID:271097938}
}

@article{Lewis2020RetrievalAugmentedGF,
  title={Retrieval-Augmented Generation for Knowledge-Intensive NLP Tasks},
  author={Patrick Lewis and Ethan Perez and Aleksandara Piktus and Fabio Petroni and Vladimir Karpukhin and Naman Goyal and Heinrich Kuttler and Mike Lewis and Wen-tau Yih and Tim Rockt{\"a}schel and Sebastian Riedel and Douwe Kiela},
  journal={ArXiv},
  year={2020},
  volume={abs/2005.11401},
  url={https://api.semanticscholar.org/CorpusID:218869575}
}

@techreport{Stouffer2023GuideTO,
  title       = {Guide to Operational Technology ({OT}) Security},
  author      = {Stouffer, Keith and Pease, Michael and Tang, CheeYee and
                 Zimmerman, Timothy and Pillitteri, Victoria and
                 Lightman, Suzanne and Hahn, Adam and Saravia, Stephanie
                 and Sherule, Aslam and Thompson, Michael},
  institution = {National Institute of Standards and Technology},
  type        = {{NIST Special Publication}},
  number      = {800-82r3},
  month       = sep,
  year        = {2023},
  doi         = {10.6028/NIST.SP.800-82r3},
  url         = {https://doi.org/10.6028/NIST.SP.800-82r3}
}

@inproceedings{Chen2024M3EmbeddingMM,
  title={M3-Embedding: Multi-Linguality, Multi-Functionality, Multi-Granularity Text Embeddings Through Self-Knowledge Distillation},
  author={Jianlv Chen and Shitao Xiao and Peitian Zhang and Kun Luo and Defu Lian and Zheng Liu},
  booktitle={Annual Meeting of the Association for Computational Linguistics},
  year={2024},
  url={https://api.semanticscholar.org/CorpusID:267413218}
}

@article{Yang2024Qwen25TR,
  title={Qwen2.5 Technical Report},
  author={Qwen An Yang and Baosong Yang and Beichen Zhang and Binyuan Hui and Bo Zheng and Bowen Yu and Chengyuan Li and Dayiheng Liu and Fei Huang and Guanting Dong and Haoran Wei and Huan Lin and Jian Yang and Jianhong Tu and Jianwei Zhang and Jianxin Yang and Jiaxin Yang and Jingren Zhou and Junyang Lin and Kai Dang and Keming Lu and Keqin Bao and Kexin Yang and Le Yu and Mei Li and Mingfeng Xue and Pei Zhang and Qin Zhu and Rui Men and Runji Lin and Tianhao Li and Tingyu Xia and Xingzhang Ren and Xuancheng Ren and Yang Fan and Yang Su and Yi-Chao Zhang and Yunyang Wan and Yuqi Liu and Zeyu Cui and Zhenru Zhang and Zihan Qiu and Shanghaoran Quan and Zekun Wang},
  journal={ArXiv},
  year={2024},
  volume={abs/2412.15115},
  url={https://api.semanticscholar.org/CorpusID:274859421}
}

@article{Wang2025InternVL35AO,
  title={InternVL3.5: Advancing Open-Source Multimodal Models in Versatility, Reasoning, and Efficiency},
  author={Weiyun Wang and Zhangwei Gao and Lixin Gu and Hengjun Pu and Long Cui and Xingguang Wei and Zhaoyang Liu and Linglin Jing and Shenglong Ye and Jie Shao and Zhaokai Wang and Zhe Chen and Hongjie Zhang and Ganlin Yang and Haomin Wang and Qi Wei and Jinhui Yin and Wenhao Li and Erfei Cui and Guanzhou Chen and Zichen Ding and Changyao Tian and Zhenyu Wu and Jingjing Xie and Zehao Li and Bowen Yang and Yuchen Duan and Xuehui Wang and Haoran Hao and Songze Li and Xiangyu Zhao and Haodong Duan and Nianchen Deng and Bin Fu and Yinan He and Yi Wang and Conghui He and Botian Shi and Junjun He and Ying Xiong and Han Lv and Lijun Wu and Wenqi Shao and Kai Zhang and Hui Deng and Biqing Qi and Biqing Qi and Qipeng Guo and Wenwei Zhang and Yuzhe Gu and Wanli Ouyang and Limin Wang and Min Dou and Xizhou Zhu and Tong Lu and Dahua Lin and Jifeng Dai and Bowen Zhou and Weijie Su and Kaiming Chen and Yu Qiao and Wenhai Wang and Gen Luo},
  journal={ArXiv},
  year={2025},
  volume={abs/2508.18265},
  url={https://api.semanticscholar.org/CorpusID:280710824}
}

@misc{qwen2.5-VL,
    title = {Qwen2.5-VL},
    url = {https://qwenlm.github.io/blog/qwen2.5-vl/},
    author = {Qwen Team},
    month = {January},
    year = {2025}
}

@article{comanici2025gemini,
  title={Gemini 2.5: Pushing the frontier with advanced reasoning, multimodality, long context, and next generation agentic capabilities},
  author={Comanici, Gheorghe and Bieber, Eric and Schaekermann, Mike and Pasupat, Ice and Sachdeva, Noveen and Dhillon, Inderjit and Blistein, Marcel and Ram, Ori and Zhang, Dan and Rosen, Evan and others},
  journal={arXiv preprint arXiv:2507.06261},
  year={2025}
}

@inproceedings{lantz2010network,
  title={A network in a laptop: rapid prototyping for software-defined networks},
  author={Lantz, Bob and Heller, Brandon and McKeown, Nick},
  booktitle={Proceedings of the 9th ACM SIGCOMM Workshop on Hot Topics in Networks},
  pages={1--6},
  year={2010}
}

@inproceedings{kwon2023efficient,
  title={Efficient memory management for large language model serving with pagedattention},
  author={Kwon, Woosuk and Li, Zhuohan and Zhuang, Siyuan and Sheng, Ying and Zheng, Lianmin and Yu, Cody Hao and Gonzalez, Joseph and Zhang, Hao and Stoica, Ion},
  booktitle={Proceedings of the 29th symposium on operating systems principles},
  pages={611--626},
  year={2023}
}

@article{gao2023retrieval,
  title={Retrieval-Augmented Generation for Large Language Models: A Survey},
  author={Gao, Yunfan and Xiong, Yun and Gao, Xinyu and Jia, Kangxiang and Pan, Jinliu and Bi, Yuxi and Dai, Yixin and Sun, Jiawei and Wang, Haofen},
  journal={arXiv preprint arXiv:2312.10997},
  year={2023}
}

@article{johnson2019billion,
  title={Billion-Scale Similarity Search with {GPU}s},
  author={Johnson, Jeff and Douze, Matthijs and J{\'e}gou, Herv{\'e}},
  journal={IEEE Transactions on Big Data},
  volume={7},
  number={3},
  pages={535--547},
  year={2019},
  publisher={IEEE}
}

\newpage

\appendices

\section{Topology Contract Schema}
\label{app:contract-schema}

Our system uses a JSON contract to constrain topology generation. The contract specifies the scenario, zones, nodes, node-level links, and zone-level links. Table~\ref{tab:contract-fields} summarizes the main fields, followed by a compact schema skeleton.

\begin{center}
\small
\refstepcounter{table}
\label{tab:contract-fields}
\textbf{Table~\thetable. Main fields in the topology contract.}

\vspace{0.5em}

\begin{tabular}{p{0.34\linewidth} p{0.56\linewidth}}
\hline
\textbf{Field} & \textbf{Description} \\
\hline
\texttt{schema\_version} & Version of the contract schema. \\
\texttt{template\_id} & Identifier of the topology template. \\
\texttt{title} & Human-readable title of the generated topology. \\
\texttt{description} & Brief description of the generation task. \\
\texttt{scenario} & Deployment scenario, e.g., enterprise, cloud, hospital, finance, or government. \\
\texttt{topology\_summary} & Natural-language summary of the generated topology. \\
\texttt{zones} & List of security or administrative zones. \\
\texttt{z.zone\_id} & Unique identifier of a zone. \\
\texttt{z.name} & Human-readable zone name. \\
\texttt{z.zone\_type} & Normalized zone category, such as enterprise, DMZ, data center, management, OT, IoT, or external. \\
\texttt{nodes} & List of network assets assigned to zones. \\
\texttt{n.node\_id} & Unique identifier of a node. \\
\texttt{n.name} & Human-readable node name. \\
\texttt{n.normalized\_type} & Normalized asset category, such as router, switch, firewall, VPN gateway, server, database, endpoint, or cloud service. \\
\texttt{n.zone\_id} & Identifier of the zone containing the node. \\
\texttt{links.node\_links} & Asset-level links between nodes. \\
\texttt{links.zone\_links} & Zone-level links between zones. \\
\texttt{link\_id} & Unique identifier of a link. \\
\texttt{source} / \texttt{target} & Source and target endpoints of a link. \\
\texttt{link\_type} & Normalized link category, such as routing, VPN, trunk, management, monitoring, or data flow. \\
\texttt{label} & Human-readable label for the link. \\
\hline
\end{tabular}
\end{center}

\Needspace{42\baselineskip}
\noindent\begin{minipage}{\columnwidth}
\noindent\textbf{Listing A.1. Compact JSON skeleton of the topology contract.}

\vspace{0.4em}

\begin{lstlisting}[basicstyle=\ttfamily\scriptsize,breaklines=true,frame=single]
{
  "schema_version": "2.1",
  "template_id": "...",
  "title": "...",
  "description": "...",
  "scenario": "...",
  "topology_summary": "...",
  "zones": [
    {
      "zone_id": "z1",
      "name": "...",
      "zone_type": "..."
    }
  ],
  "nodes": [
    {
      "node_id": "n1",
      "name": "...",
      "normalized_type": "...",
      "zone_id": "z1"
    }
  ],
  "links": {
    "node_links": [
      {
        "link_id": "nl1",
        "source": "n1",
        "target": "n2",
        "link_type": "...",
        "label": "..."
      }
    ],
    "zone_links": [
      {
        "link_id": "zl1",
        "source": "z1",
        "target": "z2",
        "link_type": "...",
        "label": "..."
      }
    ]
  }
}
\end{lstlisting}
\end{minipage}

\section{Rules-Check Contract Schema}
\label{app:rules-check-schema}

\rev{Our system uses a rules-check contract to record how the generated topology is evaluated against topology-visible CIS safeguards. The contract separates matched safeguards, rejected requirements, and items that require manual review. Each safeguard entry in the internal control database includes the safeguard identifier, title, IG applicability, topology evidence fields, and a concise predicate-style checking instruction. Table~\ref{tab:rules-check-fields} summarizes the main fields, followed by a compact schema skeleton.}
For compactness, we use the following abbreviations in Table~\ref{tab:rules-check-fields}: 
\texttt{mcc} denotes \texttt{matched\_cis\_candidates}, 
\texttt{rcr} denotes \texttt{rejected\_cis\_requirements}, 
\texttt{mrf} denotes \rev{\texttt{manual\_review\_flags}}, and 
\texttt{csi} denotes \texttt{cis\_safeguard\_id}.
\begin{center}
\small
\refstepcounter{table}
\label{tab:rules-check-fields}
\textbf{Table~\thetable. Main fields in the rules-check contract.}

\vspace{0.5em}

\begin{tabular}{p{0.36\linewidth} p{0.54\linewidth}}
\toprule
\textbf{Field} & \textbf{Description} \\
\midrule
\texttt{schema\_version} & Version of the rules-check contract schema. \\
\texttt{template\_id} & Identifier of the topology or checking template. \\
\texttt{model} & Model used to produce the rules-check output. \\
\texttt{mcc} & CIS safeguards matched to the generated topology and considered satisfied. \\
\texttt{mcc.csi} & Identifier of the matched CIS safeguard. \\
\texttt{mcc.title} & Human-readable title of the matched safeguard. \\
\texttt{mcc.status} & Matching result, e.g., satisfied. \\
\texttt{mcc.confidence} & Confidence score assigned to the safeguard match. \\
\texttt{rcr} & CIS safeguards rejected as not applicable or not supported by the generated topology. \\
\texttt{rcr.csi} & Identifier of the rejected CIS safeguard. \\
\texttt{rcr.title} & Human-readable title of the rejected safeguard. \\
\texttt{rcr.reject\_reason} & Explanation for rejecting the safeguard requirement. \\
\texttt{mrf} & CIS safeguards that require human inspection before final acceptance. \\
\texttt{mrf.csi} & Identifier of the safeguard requiring manual review. \\
\texttt{mrf.reason} & Explanation for why manual review is needed. \\
\bottomrule
\end{tabular}
\end{center}

\noindent\textbf{Listing A.2. Compact JSON skeleton of the rules-check contract.}

\begin{lstlisting}[basicstyle=\ttfamily\scriptsize,breaklines=true,frame=single]
[
  {
    "schema_version": "2.1",
    "template_id": "...",
    "model": "...",
    "matched_cis_candidates": [
      {
        "cis_safeguard_id": "13.4",
        "title": "...",
        "status": "satisfied",
        "confidence": 0.0
      }
    ],
    "rejected_cis_requirements": [
      {
        "cis_safeguard_id": "12.5",
        "title": "...",
        "reject_reason": "..."
      }
    ],
    "manual_review_flags": [
      {
        "cis_safeguard_id": "...",
        "reason": "..."
      }
    ]
  }
]
\end{lstlisting}

\section{XML Export Contract Schema}
\label{app:xml-export-schema}

The final stage exports the repaired topology, policies, compliance evidence, repair history, data flows, and residual risks into an XML contract. This XML representation is intended as a stable exchange format after validation and repair. Table~\ref{tab:xml-export-mapping} summarizes the mapping from the intermediate JSON contract to the XML export format, followed by a compact XML skeleton.

\begin{center}
\scriptsize
\refstepcounter{table}
\label{tab:xml-export-mapping}
\textbf{Table~\thetable. Mapping from intermediate JSON fields to XML export elements.}

\vspace{0.5em}

\begin{tabular}{p{0.34\linewidth} p{0.56\linewidth}}
\toprule
\textbf{XML Element} & \textbf{Description} \\
\midrule
\texttt{Topology} & Root element containing schema version, template identifier, title, scenario, and IG level. \\
\texttt{Description} & Human-readable description of the generated topology. \\
\texttt{Summary} & Human-readable summary of the generated topology. \\
\texttt{Domains/Domain} & Security or administrative zones mapped from \texttt{zones}. \\
\texttt{Nodes/Node} & Network assets mapped from \texttt{nodes}. \\
\texttt{Attributes/Attribute} & Optional node-level metadata, such as exposure, privilege, target status, or asset level. \\
\texttt{Links/ZoneLinks} & Zone-level links mapped from \texttt{links.zone\_links}. \\
\texttt{Links/NodeLinks} & Node-level links mapped from \texttt{links.node\_links}. \\
\texttt{Policies/Allow} & Connectivity test cases or policies expected to be allowed. \\
\texttt{Policies/Deny} & Connectivity test cases or policies expected to be denied. \\
\texttt{Compliance/Control} & CIS safeguard evaluation result with evidence and optional remediation. \\
\texttt{RepairHistory/Repair} & Record of topology repair actions performed to satisfy required controls. \\
\texttt{DataFlows/Flow} & Derived data-flow relationship and trust-boundary annotation. \\
\texttt{Risks/Risk} & Residual risk description and mitigation. \\
\bottomrule
\end{tabular}
\end{center}

\Needspace{42\baselineskip}
\noindent\begin{minipage}{\columnwidth}
\noindent\textbf{Listing A.3. Compact XML skeleton of the export contract.}
\vspace{0.4em}
\begin{lstlisting}[basicstyle=\ttfamily\scriptsize,breaklines=true,frame=single]
<Topology schema_version="2.1" template_id="..." title="..." scenario="cloud" ig_level="IG2">
  <Description>...</Description>
  <Summary>...</Summary>

  <Domains>
    <Domain id="z1" name="Zone Name" type="enterprise"/>
  </Domains>

  <Nodes>
    <Node id="n1" name="Node Name" domain="z1" domain_name="Zone Name" type="firewall" ip="">
      <Attributes>
        <Attribute name="asset_level" value=""/>
        <Attribute name="is_target" value="false"/>
        <Attribute name="is_exposed" value="false"/>
        <Attribute name="is_high_priv" value="false"/>
      </Attributes>
    </Node>
  </Nodes>

  <Links>
    <ZoneLinks>
      <Link id="zl1" source="z1" target="z2" type="routing" label=""/>
    </ZoneLinks>
    <NodeLinks>
      <Link id="nl1" source="n1" target="n2" type="data_flow" label=""/>
    </NodeLinks>
  </Links>

  <Policies>
    <Allow id="tc1" src_zone="z1" dst_zone="z2" src_node="n1" dst_node="n2" protocol="tcp" port="443">
      <Reason>...</Reason>
    </Allow>
    <Deny id="tc2" src_zone="z3" dst_zone="z1" src_node="" dst_node="" protocol="" port="">
      <Reason>...</Reason>
    </Deny>
  </Policies>

  <Compliance ig_level="IG2" all_required_satisfied="true">
    <Control id="13.4" title="Perform Traffic Filtering Between Network Segments" status="satisfied" confidence="1.0">
      <Evidence>...</Evidence>
      <Remediation></Remediation>
    </Control>
  </Compliance>

  <RepairHistory>
    <Repair iteration="1" control_id="13.4" action="add" target_type="node" target_id="n1">
      <Reason>...</Reason>
    </Repair>
  </RepairHistory>

  <DataFlows>
    <Flow id="df1" src="z1" dst="z2" protocol="" port="" trust_boundary="true">
      <Reason>...</Reason>
    </Flow>
  </DataFlows>

  <Risks>
    <Risk id="risk1" severity="medium" control_id="13.4">
      <Description>...</Description>
      <Mitigation>...</Mitigation>
    </Risk>
  </Risks>
</Topology>
\end{lstlisting}
\end{minipage}

\section{Runtime Prompts}
\label{app:runtime-prompts}

Our pipeline uses stage-specific system prompts to constrain generation, repair, verification, and test-case construction. Rather than using a single open-ended prompt, each stage assigns the model a specific role, restricts the allowed output format, and enforces additive or verification-only behavior. Table~\ref{tab:runtime-prompts} summarizes the runtime prompts used in the pipeline.

\begin{center}
\scriptsize
\refstepcounter{table}
\label{tab:runtime-prompts}
\textbf{Table~\thetable. Summary of runtime prompts in the TopoIntent pipeline.}

\vspace{0.5em}

\begin{tabular}{p{0.20\linewidth} p{0.28\linewidth} p{0.42\linewidth}}
\toprule
\textbf{Prompt} & \textbf{Role} & \textbf{Purpose} \\
\midrule
\texttt{Stage1} & Network security topology analyst & Extract structured fields from natural-language intent and produce an initial topology template with scenario, zones, nodes, and empty links. \\

\texttt{Stage3-Fusion} & Network security topology architect & Add minimal missing topology elements using reference templates while preserving user-specified zones, nodes, and intent requirements. \\

\texttt{Stage3-Repair} & Network security compliance engineer & Add minimal zones, nodes, or links needed to support target CIS safeguards without claiming that controls are fully verified. \\

\texttt{Stage4-Verify} & CIS Controls topology compliance auditor & Verify whether the topology satisfies each required CIS safeguard using only explicit or strongly implied topology evidence. \\

\texttt{Stage4-Repair} & Network security topology repair engineer & Propose compact additive repairs for unsatisfied or ambiguous CIS findings while preserving structural integrity. \\

\texttt{Stage4-TestCase} & Network security test engineer & Generate final executable node-to-node connectivity tests after topology verification and repair. \\
\bottomrule
\end{tabular}
\end{center}

\section{Dataset Construction Prompts}
\label{app:dataset-construction-prompts}

Before running the TopoIntent pipeline, we construct and curate the dataset through four prompt-guided preprocessing steps: topology extraction, CIS topology-relevance screening, judge-based consolidation, and synthetic intent generation. Table~\ref{tab:data-construction-prompts} summarizes these prompts and their roles.

\begin{center}
\scriptsize
\refstepcounter{table}
\label{tab:data-construction-prompts}
\textbf{Table~\thetable. Summary of dataset construction prompts.}

\vspace{0.5em}

\begin{tabular}{p{0.25\linewidth} p{0.25\linewidth} p{0.40\linewidth}}
\toprule
\textbf{Prompt} & \textbf{Role} & \textbf{Purpose} \\
\midrule
\texttt{Topology Extraction} & Extraction engine & Extract only visible zones, nodes, node links, and zone links from topology diagrams. The prompt forbids hidden devices, hidden policies, inferred controls, IP addresses, and CIS mapping. \\

\texttt{CIS Rules Check} & \rev{Rules classifier} & Classify CIS safeguards as topology-related or not. A safeguard is marked topology-related only when it directly affects zones, boundaries, links, devices, remote access paths, data flows, exposed assets, or network monitoring points. \\

\texttt{As-Judge} & Dataset judge & Consolidate multiple model outputs into a final topology template and final CIS decision. The judge uses the original image as the source of truth and classifies each CIS candidate as matched, rejected, or requiring manual review. \\

\texttt{Intent Generation} & User-intent generator & Generate realistic natural-language user requirements from finalized topology ground truth and the original image. The prompt requires moderately underspecified intents and forbids internal IDs, schema names, CIS IDs, and unsupported additions. \\
\bottomrule
\end{tabular}
\end{center}

\section{Topology-Visible CIS Controls v8.1.2 Safeguards}
\label{app:cis-safeguards}

\begin{table}[!h]
\centering
\caption{Topology-Visible CIS Controls v8.1.2 Safeguards}
\label{tab:topo-visible-cis}
\begin{tabular}{llccc}
\toprule
\textbf{ID} & \textbf{Title} & \textbf{IG1} & \textbf{IG2} & \textbf{IG3} \\
\midrule
1.1   & Establish and Maintain Detailed Enterprise Asset Inventory & \checkmark & \checkmark & \checkmark \\
3.8   & Document Data Flows & & \checkmark & \checkmark \\
3.12  & Segment Data Processing and Storage Based on Sensitivity & & \checkmark & \checkmark \\
4.4   & Implement and Manage a Firewall on Servers & \checkmark & \checkmark & \checkmark \\
4.9   & Configure Trusted DNS Servers on Enterprise Assets & & \checkmark & \checkmark \\
6.4   & Require MFA for Remote Network Access & \checkmark & \checkmark & \checkmark \\
9.2   & Use DNS Filtering Services & \checkmark & \checkmark & \checkmark \\
9.3   & Maintain and Enforce Network-Based URL Filters & & \checkmark & \checkmark \\
12.2  & Establish and Maintain a Secure Network Architecture & & \checkmark & \checkmark \\
12.3  & Securely Manage Network Infrastructure & & \checkmark & \checkmark \\
12.5  & Centralize Network Authentication, Authorization, and Auditing (AAA) & & \checkmark & \checkmark \\
12.7  & Ensure Remote Devices Utilize a VPN and are Connecting to an Enterprise’s AAA Infrastructure & & \checkmark & \checkmark \\
12.8  & Establish and Maintain Dedicated Computing Resources for All Administrative Work & & & \checkmark \\
13.3  & Deploy a Network Intrusion Detection Solution & & \checkmark & \checkmark \\
13.4  & Perform Traffic Filtering Between Network Segments & & \checkmark & \checkmark \\
13.5  & Manage Access Control for Remote Assets & & \checkmark & \checkmark \\
13.6  & Collect Network Traffic Flow Logs & & \checkmark & \checkmark \\
13.8  & Deploy a Network Intrusion Prevention Solution & & & \checkmark \\
13.9  & Deploy Port-Level Access Control & & & \checkmark \\
13.10 & Perform Application Layer Filtering & & & \checkmark \\
16.8  & Separate Production and Non-Production Systems & & \checkmark & \checkmark \\
18.2  & Perform Periodic External Penetration Tests & & \checkmark & \checkmark \\
\bottomrule
\end{tabular}
\end{table}

\end{document}